%% file: main.tex
\documentclass{article}

    \usepackage[final]{neurips_2021}

\usepackage[utf8]{inputenc} 
\usepackage[T1]{fontenc}    
\usepackage{hyperref}       
\usepackage{url}            
\usepackage{booktabs}       
\usepackage{amsfonts}       
\usepackage{nicefrac}       
\usepackage{microtype}      
\usepackage{xcolor}         
\usepackage{amsmath}
\usepackage{amssymb}
\usepackage{mathtools}
\usepackage{amsthm}
\usepackage{multirow}
\usepackage{mathtools} 
\usepackage{subfig}
\input{math_commands.tex}

\usepackage{natbib} 
\usepackage[ruled,vlined]{algorithm2e}
\usepackage[capitalize]{cleveref}
\usepackage{dsfont}

\title{Molecule Design by Latent Prompt Transformer}

\author{%
  Deqian Kong \\
  UCLA\\
  \And
  Yuhao Huang \\
  Xi'an Jiaotong Univerisity \\
  \And
  Jianwen Xie \\
  BioMap \\
  \And
  Ying Nian Wu \\
  UCLA 
}

\begin{document}

\maketitle

\begin{abstract}
This paper proposes a latent prompt Transformer model for solving challenging optimization problems such as molecule design, where the goal is to find molecules with optimal values of a target chemical or biological property that can be computed by an existing software. Our proposed model consists of three components. (1) A latent vector whose prior distribution is modeled by a Unet transformation of a Gaussian white noise vector. (2) A molecule generation model that generates the string-based representation of molecule conditional on the latent vector in (1). We adopt the causal Transformer model that takes the latent vector in (1) as prompt. (3) A property prediction model that predicts the value of the target property of a molecule based on a non-linear regression on the latent vector in (1). We call the proposed model the latent prompt Transformer model. After initial training of the model on existing molecules and their property values, we then gradually shift the model distribution towards the region that supports desired values of the target property for the purpose of molecule design. Our experiments show that our proposed model achieves state of the art performances on several benchmark molecule design tasks. 
\end{abstract}

\section{Introduction}
In drug discovery, identifying or designing molecules with specific pharmacological or chemical attributes, such as enhanced drug-likeness or high binding affinity to target proteins, is of paramount importance. However, navigating the vast space of potential drug-like molecules presents a daunting challenge. 

To address this challenge, several contemporary research avenues have emerged. One prominent approach involves the application of latent space generative models. This approach strives to translate the discrete molecule graph into a more manageable continuous latent vector. Once translated, molecular properties can be optimized within the continuous latent space utilizing various strategies~\citep{gomez2018automatic, kusner2017grammar, jin2018junction,eckmann2022limo,kong23a}. Another avenue of research is more direct, employing reinforcement learning algorithms to fine-tune molecular attributes directly within the molecule graph space~\citep{you2018graph,de2018molgan,zhou2019optimization,shi2020graphaf,luo2021graphdf}. Diverse alternative methodologies have also gained traction, such as genetic algorithms~\citep{nigam2020augmenting}, particle-swarm strategies~\citep{winter2019efficient} and scaffolding tree~\citep{fu2021differentiable}.

In this paper, we propose a novel latent prompt Transformer model for molecule design. As in most existing work on molecule design, we assume that the value of a property of interest of a given molecule can be obtained by querying an existing software such as RDKit~\citep{landrum2013rdkit} and AutoDock-GPU~\citep{santos2021accelerating}. Thus in this paper, we solely focus on optimizing the value of the target property. In our paper, we work with string-based representation of molecules, such as the commonly used SMILES~\citep{weininger1988smiles} and the more recently proposed SELFIES~\citep{krenn2020self} and its variant~\citep{cheng2023group}.

Our proposed model belongs to the latent space generative modeling approach mentioned above.  Our model consists of three components. (1) A latent vector whose prior distribution is modeled by a Unet transformation of a Gaussian white noise vector. (2) A molecule generation model that generates the string-based representation of molecule given the latent vector in (1). We adopt the causal Transformer model that takes the latent vector in (1) as prompt. (3) A property prediction model that predicts the value of the target property of a molecule based on a non-linear regression on the latent vector in (1). We call the proposed model the latent prompt Transformer model. 

After initial training of the model on existing molecules and their property values, we then gradually shift the model distribution towards the region that supports desired values of the target property for the purpose of molecule design. Our experiments show that our proposed model achieves state of the art performances on several benchmark molecule design tasks. 

The contributions of our paper are as follows. (1) We propose a novel latent prompt Transformer model for modeling the joint distribution of the molecules and their values of target property. (2) We develop the approximate maximum likelihood learning algorithm to fit the model to the training molecules and their properties. We also employ a gradual distribution shifting algorithm that shifts our model distribution towards the region that supports desired values of target property. (3) We conduct experiments on single-objective and multi-objective molecule design and our experiments achieve new state of the arts on various benchmark tasks. 

\section{Method}

\subsection{Latent Prompt Transformer}

\begin{figure*}[h]
\centering
\includegraphics[width=.6\linewidth]{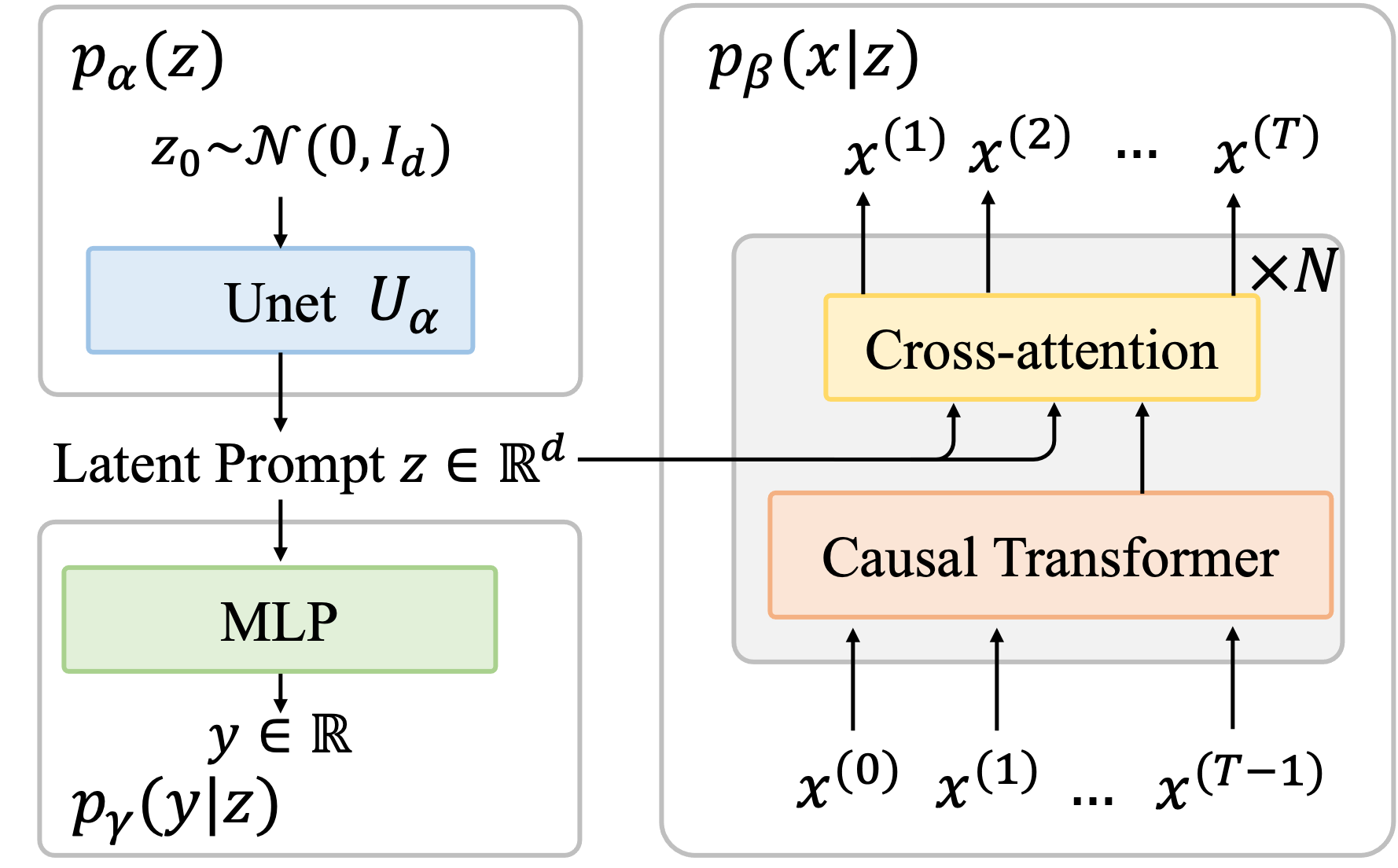}
\caption{Latent Prompt Transformer. $x$ is the string-based representation of molecule. $y$ is the value of a target property. $z$ is the latent vector. $z_0 \sim {\cal N}(0, I_d)$. (1) The prior distribution of $z$ is modeled by a Unet transformation of $z_0$, i.e., $z = U_\alpha(z_0)$. Given $z$, $x$ and $y$ are independent. (2) $p_\beta(x|z)$ is the generation model, parametrized by a causal Transformer with $z$ serving as the prompt. (3) $p_\gamma(y|z)$ is the property prediction model, which is a non-linear regression on $z$ parametrized by a multi-layer perceptron (MLP). }
\label{fig:LPT}
\end{figure*}

Our model is illustrated by \cref{fig:LPT}.  Suppose $x = (x^{(1)}, ..., x^{(t)},..., x^{(T)})$ is a molecule string in SELFIES~\citep{krenn2020self}, $y \in \mathbb{R}$ is the value of the target property of interest, and $z \in \mathbb{R}^d$ is the latent vector. We define the following model as the latent prompt Transformer (LPT):
\begin{align}
z \sim p_{\alpha}(z), \quad [x|z] \sim p_\beta(x|z), \quad [y|z] \sim p_\gamma(y|z),
\end{align}
where $p_{\alpha}(z)$ is a prior model with parameters $\alpha$.  $z$ serves as the latent prompt of the molecule generation model $p_\beta(x|z)$ parameterized by a causal Transformer with parameter $\beta$. $p_\gamma(y|z)$ is a property prediction model with parameter $\gamma$. 

For the prior model, $p_{\alpha}(z)$ is formulated as a learnable neural transport from an uninformative prior,
\begin{align}
z = U_\alpha(z_0), \label{eq:prior}
\end{align}   
where $z_0$ is assumed to be isotropic Gaussian $z_0\sim\mathcal{N}(0,I_d)$, and $U_\alpha(\cdot)$ is parametrized by an expressive neural network such as a Unet with parameter $\alpha$.

The molecule generation model $p_{\beta}(x|z)$ is a conditional autoregressive model,
\begin{align} 
    p_\beta(x|z) = \prod_{t=1}^T p_\beta(x^{(t)}|x^{(0)}, ..., x^{(t-1)}, z) 
\end{align} 
which is parameterized by a causal Transformer with parameter $\beta$. Note that the latent vector $z$ controls every step of the autoregressive generation and it functions as a soft prompt that controls the generation of molecules.

Given a molecule $x$, let $y$ denote the value of the target property, such as drug likeliness or protein binding affinity. One can determine the estimated value of this property using open-source software such as RDKit~\citep{landrum2013rdkit} and AutoDock-GPU~\citep{santos2021accelerating}.

Given $z$, we posit that $x$ and $y$ are conditionally independent. Under this assumption, LPT defines the joint distribution
\begin{align}
p_{\theta}(x, y, z) &= p_{\alpha}(z) p_\beta(x|z) p_\gamma( y|z),
\label{eq:joint}
\end{align}
where $\theta = (\alpha, \beta, \gamma)$. We use the marginal distribution $p_\theta(x, y) = \int p_\theta(x, y, z)dz $ to approximate the data distribution  $p_\mathrm{data}(x, y)$.

For the property prediction model, we assume 
\begin{align}
   p_{\gamma}(y|z) = \frac{1}{\sqrt{2 \pi \sigma^2}} \exp\left( - \frac{1}{2\sigma^2} (y - s_\gamma(z) )^2\right), \label{eq:regression}
\end{align}
where $s_\gamma(z)$ is a small multi-layer perceptron (MLP), predicting $y$ based on the latent prompt $z$. The variance $\sigma^2$ is treated as a hyper-parameter. Given this formulation, the latent prompt $z$ is aware of the property value while generating the molecule. 

For tasks involving multi-objective design with target properties $\bm{y}=\{y_j\}_{j=1}^M$, the regression model can be extended to $p_{\gamma}(\bm{y} | z) = \prod_{j=1}^M p_{\gamma_j}(y_i|z)$, where each $p_{\gamma_j}(y_i|z)$ is parametrized as in (\ref{eq:regression}). Without much loss of generality, we shall focus on the single-objective setting in the following sections.

\subsection{Learning}
\label{sec: learning}

Suppose we observe training examples $\{(x_i, y_i), i = 1, ..., n\}$. The log-likelihood function is $L( \theta) = \sum_{i=1}^{n} \log p_\theta(x_i, y_i)$.

Since $z = U_\alpha(z_0)$, we can also write the model as 
\begin{eqnarray}
   p_\theta(x, y) = \int p_{\beta}(x|z = U_\alpha(z_0)) p_{\gamma}(y| z = U_\alpha(z_0)) p_0(z_0) dz_0,
\end{eqnarray}
where $p_0(z_0) \sim {\cal N}(0, I_d)$.
The learning gradient can be calculated according to 
\begin{align} 
   \nabla_\theta  \log p_\theta(x, y)  
   = \E_{p_\theta(z_0|x, y)} \left[\nabla_\theta(\log p_\beta(x|U_\alpha(z_0)) + \log p_\gamma(y|U_\alpha(z_0))) \right].  
\end{align}

Given an example $(x, y)$, the learning gradient for the prior model is
\begin{align} 
\label{eq:alpha}
\begin{split}
  \delta_\alpha(x, y) = \E_{p_\theta(z_0|x, y)} [\nabla_\alpha(\log p_\beta(x|U_\alpha(z_0)) + \log p_\gamma(y|U_\alpha(z_0)))]. 
  \end{split}
\end{align}
The learning gradient for the molecule generation Transformer is
\begin{align} 
\label{eq:beta}
\begin{split}
  \delta_\beta(x, y) = \E_{p_\theta(z_0|x, y)} [\nabla_\beta \log p_{\beta}(x|U_\alpha(z_0))].
\end{split}
\end{align} 
The learning gradient for the property regression model is
\begin{align} 
\label{eq:gamma}
\begin{split}
  \delta_\gamma(x, y) = \E_{p_\theta(z_0|x, y)} [\nabla_\gamma \log p_{\gamma}(y|U_\alpha(z_0))].
\end{split}
\end{align} 

Estimating expectations in \cref{eq:alpha,eq:beta,eq:gamma} requires MCMC sampling of the posterior distribution $p_\theta(z_0|x,y)$. We recruit Langevin dynamics \citep{neal2011mcmc,han2017abp}. For a target distribution $\pi(z)$, the dynamics iterates
\begin{align} 
z^{\tau+1} = z^\tau + s \nabla_z \log \pi(z^\tau) + \sqrt{2s} \epsilon^\tau, 
 \label{eq:Langevin}
\end{align}
where $\tau$ indexes the time step of the Langevin dynamics, $s$ is step size, and $\epsilon_\tau \sim {\mathcal N}(0, I_d)$ is the Gaussian white noise. $\pi(z)$ here is the posterior $p_\theta(z_0|x, y)$, and the gradient can be efficiently computed by back-propagation. 

We initialize $z_0^{\tau=0} \sim {\mathcal N}(0, I_d)$, and employ $N$ steps of Langevin dynamics (e.g. $N=15$) for approximate sampling from the posterior distribution, rendering our learning algorithm as an approximate maximum likelihood estimation. See \citep{pang2020learning,nijkamp2020learning2,xie2023tale} for a theoretical understanding of the learning algorithm based on the finite-step MCMC.

In practical applications with multiple molecular generation tasks, with each characterized by a different target property $y$, each model $p_\theta(x,y)$ may necessitate separate training. For the sake of efficiency, we adopt a two-stage training approach. In the first stage, we train the model on molecules alone while ignoring the properties by maximizing $\log p_\theta(x)=\log \int p_\theta(x,z)dz$. In the second stage, we fine-tune the model for the specific target property under consideration using \cref{eq:alpha,eq:beta,eq:gamma}. To be specific, for the first pre-training stage, the learning gradient is 
\begin{align} 
\label{eq:pretrain}
   &\nabla_\theta  \log p_\theta(x)  = \E_{p_\theta(z_0|x)} \left[\nabla_\beta \log p_{\theta}(x|U_\alpha(z_0)) \right],
\end{align}
so that for a training example $(x, y)$, the learning gradients for $\alpha$ and $\beta$ are
\begin{align} 
  \delta_\alpha(x) = \E_{p_\theta(z_0|x)}[\nabla_\alpha\log p_\beta(x|U_\alpha(z_0))], \label{eq:alpha_pt}
\end{align}
\begin{align} 
\label{eq:beta_pt}
\begin{split}
  \delta_\beta(x) = \E_{p_\theta(z_0|x)} [\nabla_\beta \log p_{\beta}(x|U_\alpha(z_0))].
\end{split}
\end{align} 

The learning algorithms for pre-training LPT and fine-tuning LPT are summarized in \cref{algo:pretraining,algo:fine-tuning} respectively. This two-stage approach is also adaptable for semi-supervised scenarios where property values might be scarce.

\begin{algorithm}
	\SetKwInOut{Input}{input} \SetKwInOut{Output}{output}
	\DontPrintSemicolon
	\Input{Learning iterations~$T$, learning rates for the prior, generation models $\{\eta_0, \eta_1\}$, initial parameters~$\theta_0 = (\alpha_0, \beta_0)$, observed examples~$\{x_i\}_{i=1}^n$, batch size~$m$, number of posterior sampling steps $N_0$, and posterior sampling step size $s_0$.}
	\Output{ $\theta_T = (\alpha_{T}, \beta_{T}, \gamma_{T})$.}
	\For{$t = 0:T-1$}{			
		1. {\bf Mini-batch}: Sample observed examples $\{ x_i \}_{i=1}^m$. \\
		2. {\bf Posterior sampling}: For each $x_i$, sample $z_{0} \sim {p}_{\theta_t}(z_0|x_i)$ using~\cref{eq:Langevin}, where the target distribution $\pi$ is ${p}_{\theta_t}(z_0|x_i)$, and $s = s_0$, $N = N_0$. \\
		3. {\bf Update prior model}: $\alpha_{t+1} = \alpha_t + \eta_0 \frac{1}{m} \sum_{i=1}^{m} [\delta_\alpha(x_i)]$ as in \cref{eq:alpha_pt}. \\
		4. {\bf Update generation model}: $\beta_{t+1} = \beta_t + \eta_1 \frac{1}{m} \sum_{i=1}^{m}[\delta_\beta(x_i)]$ as in \cref{eq:beta_pt}. 
		}
	\caption{Pre-training LPT solely on molecules}
	\label{algo:pretraining}
\end{algorithm}

\begin{algorithm}[t]
	\SetKwInOut{Input}{input} \SetKwInOut{Output}{output}
	\DontPrintSemicolon
	\Input{Learning iterations~$T$, learning rates for the prior, generation, and regression models $\{\eta_0, \eta_1, \eta_2\}$, initial parameters~$\theta_0 = (\alpha_0, \beta_0, \gamma_0)$, observed samples~$\{(x_i, y_i )\}_{i=1}^n$, batch size~$m$, number of posterior sampling steps $N_1$, and posterior sampling step size $s_1$.}
	\Output{ $\theta_T = (\alpha_{T}, \beta_{T}, \gamma_{T})$.}
	\For{$t = 0:T-1$}{			
		\smallskip
		1. {\bf Mini-batch}: Sample observed examples $\{ (x_i, y_i) \}_{i=1}^m$. \\
		2. {\bf Posterior sampling}: For each $(x_i, y_i)$, sample $z_0 \sim {p}_{\theta_t}(z_0|x_i, y_i)$ using~\cref{eq:Langevin}, where the target distribution $\pi$ is  ${p}_{\theta_t}(z_0|x_i, y_i)$, and $s = s_1$, $N = N_1$. \\
		3. {\bf Update prior model}: $\alpha_{t+1} = \alpha_t + \eta_0 \frac{1}{m} \sum_{i=1}^{m} [\delta_\alpha(x_i,y_i)]$ as in \cref{eq:alpha}. \\
		4. {\bf Update generation model}: $\beta_{t+1} = \beta_t + \eta_1 \frac{1}{m} \sum_{i=1}^{m}[\delta_\beta(x_i,y_i)]$ as in \cref{eq:beta}. \\ 
		5. {\bf Update regression model}: $\gamma_{t+1} = \gamma_t + \eta_2 \frac{1}{m} \sum_{i=1}^{m}[\delta_\gamma(x_i,y_i)]$ as in \cref{eq:gamma}.
		}
	\caption{Fine-tuning or learning LPT on both molecules and their properties}
	\label{algo:fine-tuning}
\end{algorithm}

\subsection{Initial Training and Conditioned Generation}
\label{initial}

We can first pre-train the model on a dataset of existing molecules, such as ZINC~\citep{irwin2012zinc} using \cref{algo:pretraining} . Given a target property, we can then fine-tune the model using \cref{algo:fine-tuning}, where for each molecule $x$ in the training dataset, we can obtain the corresponding $y$ by querying an existing software that estimates the value of the target property, such as RDKit~\citep{landrum2013rdkit} and AutoDock-GPU~\citep{santos2021accelerating}. In this paper, we treat the values produced by the software as the ground-truth values. 

Given a trained model, we can generate a molecule $x$ conditional on a given value $y$ of the target property by sampling from $p_\theta(x|y)$. The sampling can be accomplished by the following two steps. Step 1: sample $z \sim p_\theta(z|y)$, and Step 2: sample $x \sim p_\beta(x|z)$. To accomplish Step 1, we can first sample $z_0 \sim p_\theta(z_0|y) \propto p_0(z_0) p_\gamma(y \mid z = U_\alpha(z_0))$ by Langevin dynamics, and then let $z = U_\alpha(z_0)$.

\subsection{Gradual Distribution Shifting}
\label{shifting}

\begin{algorithm}[h]
	\SetKwInOut{Input}{input} \SetKwInOut{Output}{output}
	\DontPrintSemicolon
	\Input{Shift iterations~$T$, initial pre-trained parameters $\theta_0 = (\alpha_0, \beta_0, \gamma_0)$, initial samples~$\mathcal{D}^0=\{(x_i^0, y_i^0, z_i^0) \}_{i=1}^n$ from the data distribution boundary, shift magnitude $\Delta_y$, a score function $S(x)$ and $m$ generated samples in each iteration.}
	\Output{ $\{(x_i^T, y_i^T)\}_{i=1}^n$.}
	\For{$t = 1:T$}{			
		\smallskip
		1. {\bf Dataset Creation}:  \\
  Generate $\{z_i^{t+1}, x_i^{t+1}\}_{i=1}^m$ such that $z_i^{t+1}\sim p_{\theta_t}(z_0|y=y^t+\Delta_y)$ and $x_i^{t+1}\sim p_{\beta_t}(x|z^{t+1})$. \\
  Annotate $\{y_i^{t+1}=S(x_i^{t+1})\}_{i=1}^m$. \\Create $\mathcal{G}^{t+1}=\{x_i^{t+1},y_i^{t+1},z_i^{t+1}\}_{i=1}^m\cup \mathcal{D}^{t}=\{x_i^{t+1},y_i^{t+1},z_i^{t+1}\}_{i=1}^{n+m}$. \\ 
		2. {\bf Model Shift}: \\
  Rank $\mathcal{G}^{t+1}$ based on target property $y^{t+1}$ yielding $\mathcal{G}^{t+1}=\{x_{(i)}^{t+1},y_{(i)}^{t+1},z_{(i)}^{t+1}\}_{i=1}^{n+m}$ where $y_{(1)}^{t+1}\ge y_{(2)}^{t+1}\ge\cdots\ge y_{(m+n)}^{t+1}$.\\
  Create $\mathcal{D}^{t+1}=\{x_{(i)}^{t+1},y_{(i)}^{t+1},z_{(i)}^{t+1}\}_{i=1}^{n}$.\\
  Update $\theta_{t+1}=\underset{\theta}{\text{argmax }}\E_{(x,y)\sim\mathcal{D}^{t+1}}[\log p_\theta(x,y)]
$ using \cref{algo:fine-tuning}.
		}
	\caption{Sampling with Gradual Distribution Shifting (SGDS).}
	\label{algo:sgds}
\end{algorithm}

Given an initially trained model, for the purpose of molecule design, it is tempting to set $y$ at a desired value $y^*$, and then generate $x \sim p_\theta(x|y^*)$. The problem is that $y^*$ may be out of the range of the learned distribution $p_\theta(x, y, z)$ or more specifically its marginal distribution $p_\theta(y)$. As a result, sampling from $p_\theta(x|y^*)$ amounts to out-of-distribution (OOD) extrapolation, which can be unreliable. 

The \textit{sampling with gradual distribution shifting} (SGDS) algorithm~\citep{kong23a} was proposed to address the above problem. In this algorithm, we can maintain a top-$n$ {\em shifting dataset} $\mathcal{D}^t=\{x_i^t,y_i^t,z_i^t\}_{i=1}^n$, where $t$ denotes the iteration in the SGDS algorithm. To initialize at $t = 0$, we obtain $\mathcal{D}^0$ by selecting the top-$n$ molecules from the initial training dataset such as ZINC~\citep{irwin2012zinc} by ranking them based on their values of target property. That is, $\mathcal{D}^0$ is selected at the boundary of the initial training set. In each iteration of SGDS, we incrementally increase the values of the properties in the top-$n$ shifting dataset, and then generate new molecules conditional on the increased values. Because the incrementally increased values are expected to be close to the current model distribution, the conditional generation is expected to be reliable. Nonetheless, we still query the software to obtain the ground-truth values of the target property for the newly generated molecules. We then update the top-$n$ shifting dataset by ranking the molecules in the current shifting dataset as well as the newly generated molecules based on the ground-truth values of the property. For this new top-$n$ shifting dataset, we then re-learn our model, in order for the model to catch up with the shifting data, so that further incremental shifting and generation can still be reliable. More specifically, each iteration of SGDS consists of the following steps: 

(1) Generate $m$ new molecules $\{x_i^{t+1}\}_{i=1}^m\sim p_{\theta_t}(x|y=\Tilde{y}^t)$. Here, $\Tilde{y}^t=y^t+\Delta_y$, where $\Delta_y$ is a small increment, and $y^t$ is the ground-truth property value of a molecule randomly sampled from the current shifting dataset $\mathcal{D}^t$. To accomplish generation, we first sample $z^{t+1} \sim p_{\theta_t}(z|\Tilde{y}^t)$ and then use the generation model to get $x^{t+1}\sim p_{\beta_t}(x|z^{t+1})$. To sample $z^{t+1}$, we need to run finite-step Langevin dynamics to sample from $p_{\theta_t}(z_0|\Tilde{y}^t)$. This Langevin dynamics is initialized from the corresponding $z_0$ we keep in the previous shifting iteration, from which we run a very small number of Langevin steps (typically 2 steps). 

(2) Annotate the generated molecules using the software (e.g., AutoDock-GPU or RDKit), which is a black-box score or reward function $S(x)$, i.e. $\{y_i^{t+1}=S(x_i^{t+1})\}_{i=1}^m$. Different from \citep{kong23a}, we do not assume $m=n$. The updated dataset, \(\mathcal{G}^{t+1}\), combines the newly generated samples with the previous dataset: $\mathcal{G}^{t+1}=\{x_i^{t+1},y_i^{t+1},z_i^{t+1}\}_{i=1}^m\cup \mathcal{D}^{t}$, amounting to $n+m$ samples. For simplicity in notation, we write $\mathcal{G}^{t+1}=\{x_i^{t+1},y_i^{t+1},z_i^{t+1}\}_{i=1}^{n+m}$. 

(3) Rank $n+m$ samples based on target property value $y^{t+1}$. This yields $\mathcal{G}^{t+1}=\{x_{(i)}^{t+1},y_{(i)}^{t+1},z_{(i)}^{t+1}\}_{i=1}^{n+m}$ where $y_{(1)}^{t+1}\ge y_{(2)}^{t+1}\ge\cdots\ge y_{(m+n)}^{t+1}$. From this, the top-$n$ samples are kept to create a new shifting dataset: $\mathcal{D}^{t+1}=\{x_{(i)}^{t+1},y_{(i)}^{t+1},z_{(i)}^{t+1}\}_{i=1}^{n}$. Additional heuristic constraints can be applied during this selection. For instance, instead of ranking by $y_i^{t+1}$, we might rank by $y_i^{t+1}\mathds{1}_{S^\prime(x_i)>s}$ where $S^\prime(x_i)$ is another score function, $s$ is the desired threshold and $\mathds{1}$ is the indicator function.

(4) Update the model parameter $\theta^{t+1}$ by learning from the new shifting dataset $\mathcal{D}^{t+1}$ using \cref{algo:fine-tuning}.

The integration of steps (1) and (2) constitutes the primary phase of the SGDS algorithm: dataset creation.
Subsequently, the combination of steps (3) and (4) forms the second phase: model shift. 

While \citep{kong23a} proposes shifting a latent space energy-based model, our aim here is to apply SGDS for shifting our LPT.

\section{Experiments}
We demonstrate our proposed molecule design approach for both single and multi-objective settings.

\subsection{Experiment Setup}
\paragraph{Dataset.} For molecule property optimization tasks, we use ZINC~\citep{irwin2012zinc} with $250$k molecules. RDKit is used to calculate penalized logP, drug-likeliness (QED) and synthetic accessibility score (SA), and we use docking scores from AutoDock-GPU to approximate the binding affinity to two protein targets, human
estrogen receptor (ESR1) and human peroxisomal acetyl-CoA acyl transferase 1 (ACAA1).

\paragraph{Model Architectures.} 
As shown in \cref{fig:LPT}, the prior model is underpinned by Unet1D, assuming $4$ latent vectors for $z$ with each sized at $256$. The molecule generation model leverages a 3-layer causal Transformer complemented by a cross-attention layer. It has an embedding size of $256$ and uses a maximum SELFIES sequence length of $73$. The property regression model utilizes a 3-layer MLP, accepting inputs sized at $1024$ ($256\times4$). The total number of parameters for our LPT is $4.33$M.

\paragraph{Training Details.}
We adopt a two-step training approach for LPT. Initially, we pre-train on molecules alone for 30 epochs using \cref{algo:pretraining}, with a learning rate ranging between $7.5\times10^{-4}$ and $7.5\times10^{-5}$ via cosine scheduling. Subsequently, we finetune for 10 epochs on both molecules and their properties using \cref{algo:fine-tuning}, adjusting the learning rate between $3\times 10^{-4}$ and $7.5\times 10^{-5}$.
For SGDS process, the total shifting iterations is $25$ and the number of new generated samples is set at $2500$ for each iteration, with total $62.5$k queries of the software. 
We use the AdamW optimizer~\citep{loshchilov2017decoupled} with $0.1$ weight decay for all the above learning processes. Pre-training LPT, fine-tuning LPT and shifting LPT take around $20$, $10$ and $12$ hours respectively on a single NVIDIA A6000.

\subsection{Binding Affinity Maximization}
ESR1 and ACAA1 are human proteins. Our goal is to design ligands with optimal binding affinities to these proteins. While ESR1 has many known binders, SGDS disregards binder-specific data. Binding affinity is measured by the estimated dissociation constants, \(\mathrm{K_D}\), which can be approximated by AutoDock-GPU's docking scores. A lower \(\mathrm{K_D}\) indicates stronger binding. Our model excels in the single objective ESR1 and ACAA1 binding affinity maximization tasks, as highlighted in \cref{tab:single_bio}. Compared to other state-of-the-arts, it consistently samples high-affinity molecules in the shifting trajectories. Specifically, our latent prompt Transformer outperforms LEBM-SGDS~\citep{kong23a}, showcasing its robust modeling capabilities. Additionally, unlike LEBM, our LPT can readily scale its complexity of prior model and generative Transformer, making it more adaptable to larger datasets and training scenarios. 

For multi-objective optimization tasks, we consider maximizing binding affinity, QED and minimizing SA. Meanwhile, we also recruit heuristics to set a threshold to select more probable molecule. In \cref{algo:sgds}, we exclude molecules with QED smaller than $0.4$ and SA larger than $5.5$.  Results in \cref{tab:multi} show that LPT is able to get comparable QED and SA to LEBM while getting much higher binding affinities, which demonstrates its superior modeling capability. Generated molecules can be found in Appendix.

\begin{figure*}[t!]
  \centering
  \includegraphics[width=.8\textwidth,trim=140 0 110 40,clip]{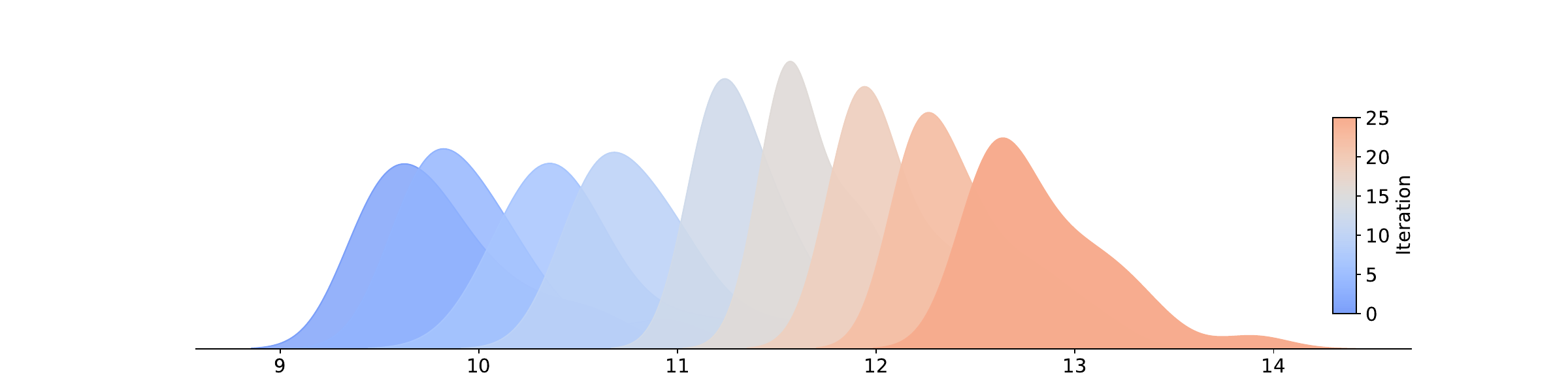}
  \caption{Distribution shift of ACAA1 binding affinity across optimization iterations. For each shift iteration, we plot the densities of property values estimated from AutoDock-GPU.}
  \label{fig:shift_illustrate}
\end{figure*}

\begin{table}[h]
\caption{Single-objective binding affinity optimization. Report top-3 lowest $\mathrm{K_D}$ (in nanomoles/liter) found by each model. Baseline results obtained from~\citep{eckmann2022limo,kong23a}.}
\label{tab:single_bio}
\centering
\resizebox{0.65\linewidth}{!}{%
\begin{tabular}{lcccccc}
\toprule
\multicolumn{1}{c}{\multirow{2}{*}{\bf Method}} & \multicolumn{3}{c}{\bf ESR1 $\mathrm{K_D}$ $(\downarrow)$} & \multicolumn{3}{c}{\bf ACAA1 $\mathrm{K_D}$ $(\downarrow)$}\\
\multicolumn{1}{c}{}                                           & 1st           & 2rd           & 3rd          & 1st           & 2rd           & 3rd                                                   \\ \midrule
GCPN                                                            & 6.4           & 6.6           & 8.5          & 75            & 83            & 84                                                    \\
MolDQN                                                          & 373           & 588           & 1062         & 240           & 337           & 608                                                   \\
MARS                                                            & 25            & 47            & 51           & 370           & 520           & 590                                                  \\
GraphDF                                                         & 17            & 64            & 69           & 163           & 203           & 236                                                  \\
LIMO                                                            & 0.72          & 0.89          & 1.4          & 37            & 37            & 41                                                    \\ 

LEBM-SGDS                                                          &  0.03              & 0.03              &  0.04            &  0.11              &  0.11              &  0.12  \\
\midrule
\textbf{LPT-SGDS}                                                            & \bf 0.004              &\bf 0.005              &\bf  0.014            & \bf 0.037              & \bf 0.046              & \bf 0.084  
\end{tabular}}
\end{table}

\begin{table}[h]
\caption{Muli-objective optimization for both ESR1 and ACAA1. Report Top-2 average scores of $\mathrm{K_D}$ (in nmol/L), QED and SA. Baseline results obtained from~\citep{eckmann2022limo,kong23a}.}
\label{tab:multi}
\centering
\resizebox{0.7\linewidth}{!}{
\begin{tabular}{lcccccc}
\toprule
\multicolumn{1}{c}{\multirow{2}{*}{\bf Ligand}} & \multicolumn{3}{c}{\bf ESR1}  & \multicolumn{3}{c}{\bf ACAA1}                                     \\
\multicolumn{1}{c}{}                        &$\mathrm{K_D}$ $\downarrow$            &\bf QED $\uparrow$                      &\bf SA $\downarrow$  &$\mathrm{K_D}$ $\downarrow$            &\bf QED $\uparrow$                      &\bf SA $\downarrow$                    \\\midrule\multicolumn{1}{l}{Tamoxifen}               & 87                   & 0.45                     &   2.0      & $-$ & $-$ & $-$               \\
\multicolumn{1}{l}{Raloxifene}              & $7.9\times10^6$      & 0.32 & 2.4  & $-$ & $-$ & $-$\\ \midrule
\multicolumn{1}{l}{GCPN $\texttt{1}^\texttt{st}$}                & 810                  & 0.43                     & 4.2   & 8500                 &  0.69                     & 4.2                  \\
\multicolumn{1}{l}{GCPN $\texttt{2}^\texttt{nd}$}                & $27000$      & 0.80                     & 3.7   & 8500                 & 0.54                     & 4.3                  \\
\multicolumn{1}{l}{LIMO $\texttt{1}^\texttt{st}$}                & 4.6                  & 0.43                     & 4.8   & 28                   & 0.57                     & 5.5                  \\
\multicolumn{1}{l}{LIMO $\texttt{2}^\texttt{nd}$}                & 2.8                  & 0.64                     & 4.9   & 31                   & 0.44                     & 4.9                  \\
\multicolumn{1}{l}{{LEBM-SGDS} $\texttt{1}^\texttt{st}$}                & 0.36                 & 0.44                     & 3.99   &  4.55             & 0.56                     & 4.07                 \\
\multicolumn{1}{l}{{LEBM-SGDS} $\texttt{2}^\texttt{nd}$}                & 1.28                 &  0.44                     & 3.86     & 5.67             & 0.60                     & 4.58        \\\midrule
\multicolumn{1}{l}{\textbf{LPT-SGDS} $\texttt{1}^\texttt{st}$}                & \bf 0.04             & 0.58                     &   \bf 3.46     &\bf 0.18                 & 0.50                     & 4.85               \\
\multicolumn{1}{l}{\textbf{LPT-SGDS} $\texttt{2}^\texttt{nd}$}                & \bf 0.05             & 0.46                     & 3.24   & 0.21                 &  \bf 0.61                     & \bf 4.18     
\end{tabular}}
\end{table}

\section{Related Work}
Our model is based on \citep{kong23a}. The difference are as follows. (1) While \citep{kong23a} used the LSTM model for molecule generation, we adopt a more expressive causal Transformer model for generation, with the latent vector serving as latent prompt. (2) While \citep{kong23a} used a latent space energy-based model for the prior of the latent vector, we assume that the latent $z$ is generated by a Unet transformation of a Gaussian white noise vector. This enables us to avoid the Langevin dynamics for prior sampling in  learning, thus simplifies the learning algorithm. (3) We obtain much stronger experimental results, surpassing \citep{kong23a} and achieving new state of the art performances. 

Compared to existing latent space generative models~\citep{gomez2018automatic, kusner2017grammar, jin2018junction,eckmann2022limo}, we assume a learnable prior model so that our model can adeptly catch up with the shifting dataset in the optimization process. 

Compared to population-based methods such as genetic algorithms~\citep{nigam2020augmenting} and particle-swarm algorithms~\citep{winter2019efficient}, our method does not only maintain a shifting dataset (which can be considered a small population), but also a shifting model to fit the dataset, so that we can generate new molecules from the model. The model itself is virtually an infinite population because it can generate infinitely many new samples.

\section{Conclusion}

This paper proposes a latent prompt Transformer model for molecule design. We assume the solution can be represented by a sequence of tokens. We employ a latent prompt that generates the sequence via a causal Transformer model and predicts the value of the target property via a regression model. We  develop the approximate maximum likelihood learning algorithm and we employ the gradual distribution shifting algorithm for optimization with learning in the loop. Our proposed method achieves new state of the art on several benchmark tasks on molecule design. 

Our model and method can be applied to on-line black-box optimization problem in general, and the Transformer model can be replaced by other conditional generation models if the solution is not in the form of a sequence of tokens. In our future work, we shall explore applying our method to other challenging optimization problems in science and engineering.

\section*{Acknowledgement}

Y. N. Wu was partially supported by NSF DMS-2015577 and a gift fund from Amazon.
\bibliography{molecule_design}

\section*{Appendix}
We display molecules generated by LPT as they evolve throughout the shifting trajectories.
\begin{figure*}[h]
  \centering
  \includegraphics[width=.67\textwidth]{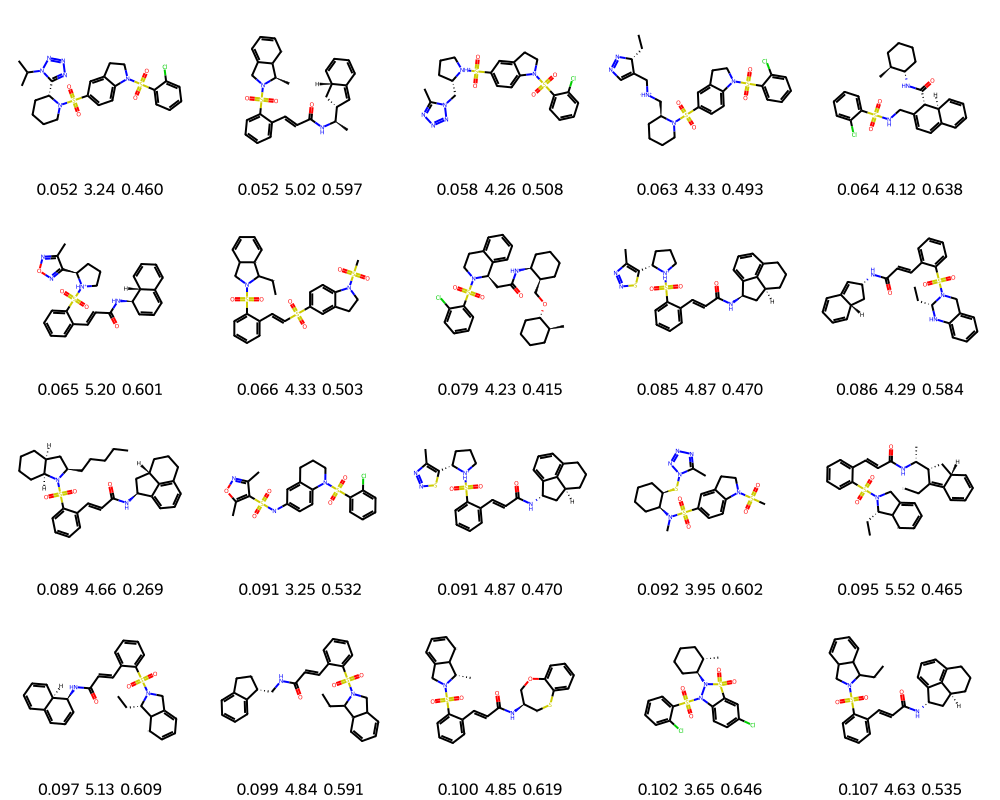}
  \caption{Molecules produced during the multi-objective optimization for ESR1. The legends denote $\mathrm{K_D}\downarrow$, SA$\downarrow$ and QED$\uparrow$.}
\end{figure*}

\begin{figure*}[h]
  \centering
  \includegraphics[width=.67\textwidth]{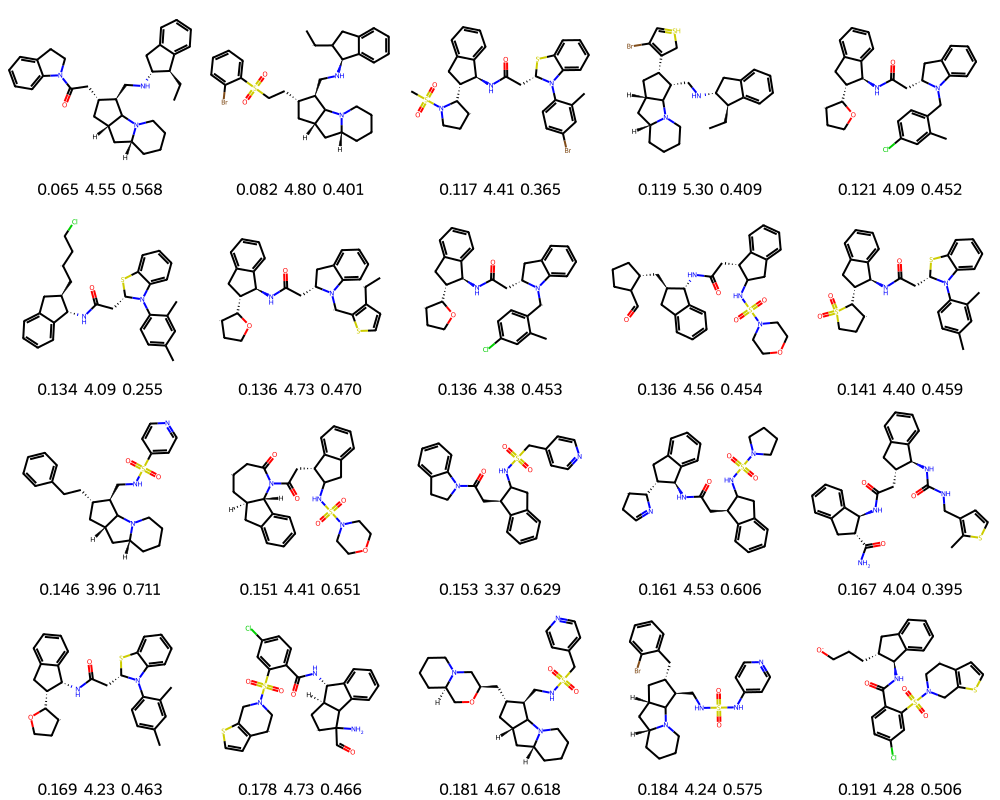}
  \caption{Molecules produced during the multi-objective optimization for ACAA1. The legends denote $\mathrm{K_D}\downarrow$, SA$\downarrow$ and QED$\uparrow$.}
\end{figure*}


\end{document}

%% file: math_commands.tex
\usepackage{amsmath,amsfonts,bm}

\def\eqref#1{equation~\ref{#1}}

\def\1{\bm{1}}

\DeclareMathAlphabet{\mathsfit}{\encodingdefault}{\sfdefault}{m}{sl}
\SetMathAlphabet{\mathsfit}{bold}{\encodingdefault}{\sfdefault}{bx}{n}

\newcommand{\E}{\mathbb{E}}




%% file: main.bbl
\begin{thebibliography}{25}
\providecommand{\natexlab}[1]{#1}
\providecommand{\url}[1]{\texttt{#1}}
\expandafter\ifx\csname urlstyle\endcsname\relax
  \providecommand{\doi}[1]{doi: #1}\else
  \providecommand{\doi}{doi: \begingroup \urlstyle{rm}\Url}\fi

\bibitem[Cheng et~al.(2023)Cheng, Cai, Miret, Malkomes, Phielipp, and Aspuru-Guzik]{cheng2023group}
Austin~H Cheng, Andy Cai, Santiago Miret, Gustavo Malkomes, Mariano Phielipp, and Al{\'a}n Aspuru-Guzik.
\newblock Group selfies: a robust fragment-based molecular string representation.
\newblock \emph{Digital Discovery}, 2023.

\bibitem[De~Cao and Kipf(2018)]{de2018molgan}
Nicola De~Cao and Thomas Kipf.
\newblock Molgan: An implicit generative model for small molecular graphs.
\newblock \emph{arXiv preprint arXiv:1805.11973}, 2018.

\bibitem[Eckmann et~al.(2022)Eckmann, Sun, Zhao, Feng, Gilson, and Yu]{eckmann2022limo}
Peter Eckmann, Kunyang Sun, Bo~Zhao, Mudong Feng, Michael~K Gilson, and Rose Yu.
\newblock Limo: Latent inceptionism for targeted molecule generation.
\newblock In \emph{International Conference on Machine Learning (ICML)}, 2022.

\bibitem[Fu et~al.(2021)Fu, Gao, Xiao, Yasonik, Coley, and Sun]{fu2021differentiable}
Tianfan Fu, Wenhao Gao, Cao Xiao, Jacob Yasonik, Connor~W Coley, and Jimeng Sun.
\newblock Differentiable scaffolding tree for molecular optimization.
\newblock \emph{arXiv preprint arXiv:2109.10469}, 2021.

\bibitem[G{\'o}mez-Bombarelli et~al.(2018)G{\'o}mez-Bombarelli, Wei, Duvenaud, Hern{\'a}ndez-Lobato, S{\'a}nchez-Lengeling, Sheberla, Aguilera-Iparraguirre, Hirzel, Adams, and Aspuru-Guzik]{gomez2018automatic}
Rafael G{\'o}mez-Bombarelli, Jennifer~N Wei, David Duvenaud, Jos{\'e}~Miguel Hern{\'a}ndez-Lobato, Benjam{\'\i}n S{\'a}nchez-Lengeling, Dennis Sheberla, Jorge Aguilera-Iparraguirre, Timothy~D Hirzel, Ryan~P Adams, and Al{\'a}n Aspuru-Guzik.
\newblock Automatic chemical design using a data-driven continuous representation of molecules.
\newblock \emph{ACS Central Science}, 4\penalty0 (2):\penalty0 268--276, 2018.

\bibitem[Han et~al.(2017)Han, Lu, Zhu, and Wu]{han2017abp}
Tian Han, Yang Lu, Song{-}Chun Zhu, and Ying~Nian Wu.
\newblock Alternating back-propagation for generator network.
\newblock In \emph{{AAAI} Conference on Artificial Intelligence (AAAI)}, pages 1976--1984, 2017.

\bibitem[Irwin et~al.(2012)Irwin, Sterling, Mysinger, Bolstad, and Coleman]{irwin2012zinc}
John~J Irwin, Teague Sterling, Michael~M Mysinger, Erin~S Bolstad, and Ryan~G Coleman.
\newblock Zinc: a free tool to discover chemistry for biology.
\newblock \emph{Journal of Chemical Information and Modeling}, 52\penalty0 (7):\penalty0 1757--1768, 2012.

\bibitem[Jin et~al.(2018)Jin, Barzilay, and Jaakkola]{jin2018junction}
Wengong Jin, Regina Barzilay, and Tommi Jaakkola.
\newblock Junction tree variational autoencoder for molecular graph generation.
\newblock In \emph{International Conference on Machine Learning (ICML)}, pages 2323--2332, 2018.

\bibitem[Kong et~al.(2023)Kong, Pang, Han, and Wu]{kong23a}
Deqian Kong, Bo~Pang, Tian Han, and Ying~Nian Wu.
\newblock Molecule design by latent space energy-based modeling and gradual distribution shifting.
\newblock In \emph{Conference on Uncertainty in Artificial Intelligence (UAI)}, volume 216, pages 1109--1120, 2023.

\bibitem[Krenn et~al.(2020)Krenn, H{\"a}se, Nigam, Friederich, and Aspuru-Guzik]{krenn2020self}
Mario Krenn, Florian H{\"a}se, AkshatKumar Nigam, Pascal Friederich, and Alan Aspuru-Guzik.
\newblock Self-referencing embedded strings (selfies): A 100\% robust molecular string representation.
\newblock \emph{Machine Learning: Science and Technology}, 1\penalty0 (4):\penalty0 045024, 2020.

\bibitem[Kusner et~al.(2017)Kusner, Paige, and Hern{\'a}ndez-Lobato]{kusner2017grammar}
Matt~J Kusner, Brooks Paige, and Jos{\'e}~Miguel Hern{\'a}ndez-Lobato.
\newblock Grammar variational autoencoder.
\newblock In \emph{International Conference on Machine Learning (ICML)}, pages 1945--1954, 2017.

\bibitem[Landrum et~al.()]{landrum2013rdkit}
Greg Landrum et~al.
\newblock Rdkit: Open-source cheminformatics.
\newblock URL \url{https://www.rdkit.org}.

\bibitem[Loshchilov and Hutter(2019)]{loshchilov2017decoupled}
Ilya Loshchilov and Frank Hutter.
\newblock Decoupled weight decay regularization.
\newblock In \emph{International Conference on Learning Representations (ICLR)}, 2019.

\bibitem[Luo et~al.(2021)Luo, Yan, and Ji]{luo2021graphdf}
Youzhi Luo, Keqiang Yan, and Shuiwang Ji.
\newblock Graphdf: A discrete flow model for molecular graph generation.
\newblock In \emph{International Conference on Machine Learning (ICML)}, pages 7192--7203, 2021.

\bibitem[Neal(2011)]{neal2011mcmc}
Radford~M Neal.
\newblock {MCMC} using hamiltonian dynamics.
\newblock \emph{Handbook of Markov Chain Monte Carlo}, 2, 2011.

\bibitem[Nigam et~al.(2020)Nigam, Friederich, Krenn, and Aspuru-Guzik]{nigam2020augmenting}
AkshatKumar Nigam, Pascal Friederich, Mario Krenn, and Al{\'a}n Aspuru-Guzik.
\newblock Augmenting genetic algorithms with deep neural networks for exploring the chemical space.
\newblock In \emph{International Conference on Learning Representations (ICLR)}, 2020.

\bibitem[Nijkamp et~al.(2020)Nijkamp, Pang, Han, Zhou, Zhu, and Wu]{nijkamp2020learning2}
Erik Nijkamp, Bo~Pang, Tian Han, Linqi Zhou, Song-Chun Zhu, and Ying~Nian Wu.
\newblock Learning multi-layer latent variable model via variational optimization of short run mcmc for approximate inference.
\newblock In \emph{European Conference on Computer Vision (ECCV)}, pages 361--378, 2020.

\bibitem[Pang et~al.(2020)Pang, Han, Nijkamp, Zhu, and Wu]{pang2020learning}
Bo~Pang, Tian Han, Erik Nijkamp, Song-Chun Zhu, and Ying~Nian Wu.
\newblock Learning latent space energy-based prior model.
\newblock In \emph{Advances in Neural Information Processing Systems (NeurIPS)}, 2020.

\bibitem[Santos-Martins et~al.(2021)Santos-Martins, Solis-Vasquez, Tillack, Sanner, Koch, and Forli]{santos2021accelerating}
Diogo Santos-Martins, Leonardo Solis-Vasquez, Andreas~F Tillack, Michel~F Sanner, Andreas Koch, and Stefano Forli.
\newblock Accelerating autodock4 with gpus and gradient-based local search.
\newblock \emph{Journal of Chemical Theory and Computation}, 17\penalty0 (2):\penalty0 1060--1073, 2021.

\bibitem[Shi et~al.(2020)Shi, Xu, Zhu, Zhang, Zhang, and Tang]{shi2020graphaf}
Chence Shi, Minkai Xu, Zhaocheng Zhu, Weinan Zhang, Ming Zhang, and Jian Tang.
\newblock Graphaf: a flow-based autoregressive model for molecular graph generation.
\newblock \emph{arXiv preprint arXiv:2001.09382}, 2020.

\bibitem[Weininger(1988)]{weininger1988smiles}
David Weininger.
\newblock Smiles, a chemical language and information system. 1. introduction to methodology and encoding rules.
\newblock \emph{Journal of Chemical Information and Computer Sciences}, 28\penalty0 (1):\penalty0 31--36, 1988.

\bibitem[Winter et~al.(2019)Winter, Montanari, Steffen, Briem, No{\'e}, and Clevert]{winter2019efficient}
Robin Winter, Floriane Montanari, Andreas Steffen, Hans Briem, Frank No{\'e}, and Djork-Arn{\'e} Clevert.
\newblock Efficient multi-objective molecular optimization in a continuous latent space.
\newblock \emph{Chemical Science}, 10\penalty0 (34):\penalty0 8016--8024, 2019.

\bibitem[Xie et~al.(2023)Xie, Zhu, Xu, Li, and Li]{xie2023tale}
Jianwen Xie, Yaxuan Zhu, Yifei Xu, Dingcheng Li, and Ping Li.
\newblock A tale of two latent flows: Learning latent space normalizing flow with short-run langevin flow for approximate inference.
\newblock In \emph{The Tenth International Conference on Learning Representations (ICLR)}, 2023.

\bibitem[You et~al.(2018)You, Liu, Ying, Pande, and Leskovec]{you2018graph}
Jiaxuan You, Bowen Liu, Zhitao Ying, Vijay Pande, and Jure Leskovec.
\newblock Graph convolutional policy network for goal-directed molecular graph generation.
\newblock In \emph{Advances in Neural Information Processing Systems (NeurIPS)}, pages 6410--6421, 2018.

\bibitem[Zhou et~al.(2019)Zhou, Kearnes, Li, Zare, and Riley]{zhou2019optimization}
Zhenpeng Zhou, Steven Kearnes, Li~Li, Richard~N Zare, and Patrick Riley.
\newblock Optimization of molecules via deep reinforcement learning.
\newblock \emph{Scientific Reports}, 9\penalty0 (1):\penalty0 1--10, 2019.

\end{thebibliography}
